\PassOptionsToPackage{numbers, compress}{natbib}
\documentclass{article}

\usepackage[pdftex]{graphicx}

\usepackage[preprint]{main}
\usepackage[utf8]{inputenc} % allow utf-8 input
\usepackage[T1]{fontenc}    % use 8-bit T1 fonts
\usepackage{hyperref}       % hyperlinks
\usepackage{url}            % simple URL typesetting
\usepackage{booktabs}       % professional-quality tables
\usepackage{amsfonts}       % blackboard math symbols
\usepackage{nicefrac}       % compact symbols for 1/2, etc.
\usepackage{microtype}      % microtypography
\usepackage{xcolor}         % colors
\usepackage{multicol}
\usepackage{multirow}
\usepackage{wrapfig}
\usepackage{algorithm}
\usepackage{algpseudocode}
\usepackage{amsmath}
\usepackage{listings}
\usepackage{color}
\usepackage{tablefootnote}

\definecolor{codegreen}{rgb}{0,0.6,0}
\definecolor{codegray}{rgb}{0.5,0.5,0.5}
\definecolor{codepurple}{rgb}{0.58,0,0.82}
\definecolor{backcolour}{rgb}{0.95,0.95,0.92}
\newcommand{\customsize}{\fontsize{4pt}{5pt}\selectfont}
\title{Investigating Video Reasoning Capability of Large Language Models with Tropes in Movies}

\author{%
Hung-Ting Su$^{1}$, Chun-Tong Chao$^{1}$, Ya-Ching Hsu$^{1}$, Xudong Lin$^{2}$, Yulei Niu$^{2}$, \\ \textbf{Hung-Yi Lee}$^{1}$, \textbf{Winston H. Hsu}$^{1,3}$ \\
$^{1}$National Taiwan University~~~~$^2$Columbia University~~~~$^3$MobileDrive Technology \\
}

\begin{document}
\maketitle

\def\datasetfull{Tropes in Movies}
\def\datasetabbr{TiM}
\def\challengeii{Abstract Perception}
\def\challengeiii{Long-range Compositional Reasoning}
\def\methodfull{Face-Enhanced Viper of Role Interactions}
\def\methodabbr{FEVoRI}
\def\promptfull{Context Query Reduction}
\def\promptabbr{ConQueR}
\def\evaluationfull{AST Based Code Dignosis}
\def\evaluationabbr{ABCD}

\begin{abstract}
Large Language Models (LLMs) have demonstrated effectiveness not only in language tasks but also in video reasoning. This paper introduces a novel dataset, \textbf{\datasetfull{} (\datasetabbr{})}, designed as a testbed for exploring two critical yet previously overlooked video reasoning skills: (1) \textbf{\challengeii{}}: understanding and tokenizing abstract concepts in videos, and (2) \textbf{\challengeiii{}}: planning and integrating intermediate reasoning steps for understanding long-range videos with numerous frames. Utilizing tropes from movie storytelling, \datasetabbr{} evaluates the reasoning capabilities of state-of-the-art LLM-based approaches. Our experiments show that current methods, including Captioner-Reasoner, Large Multimodal Model Instruction Fine-tuning, and Visual Programming, only marginally outperform a random baseline when tackling the challenges of \challengeii{} and \challengeiii{}. To address these deficiencies, we propose \methodfull{} (\methodabbr{}) and \promptfull{} (\promptabbr{}), which enhance Visual Programming by fostering role interaction awareness and progressively refining movie contexts and trope queries during reasoning processes, significantly improving performance by 15 F1 points. However, this performance still lags behind human levels (40 vs. 65 F1). 
Additionally, we introduce a new protocol to evaluate the necessity of \challengeii{} and \challengeiii{} for task resolution. This is done by analyzing the code generated through Visual Programming using an Abstract Syntax Tree (AST), thereby confirming the increased complexity of \datasetabbr{}. The dataset and code are available at: \href{https://ander1119.github.io/TiM}{https://ander1119.github.io/TiM}

%(2) \challengeiii{}: planning and integrating intermediate reasoning for the final outcome. 
%(3) \challengei{}: managing long-range videos with numerous inputs. 

  % Large Language Models (LLMs) have demonstrated their effectiveness in visual and language reasoning tasks. This paper explores three critical reasoning skills in video understanding that have been previously overlooked: (1) Extraction: managing long-range videos with numerous inputs, (2) Abstraction: understanding abstract concepts depicted, and (3) Composition: planning and integrating intermediate reasoning for the final outcome. We introduce the Tropes in Movies (TiM) dataset, utilizing tropes in movie storytelling as a testbed to investigate the reasoning capabilities of state-of-the-art LLM-based approaches. Our experiments reveal that modern LLM-based approaches, including Captioner-Reasoner, Instruction Fine-tuning, or Visual Programming, do not effectively utilize attention, abstraction, and composition reasoning skills to address the TiM dataset. We also provide comprehensive analysis to guide future research in the field of LLMs.
\end{abstract}

\section{Introduction}\label{sec:1}
\begin{figure}
  \centering
  \includegraphics[width=\textwidth]{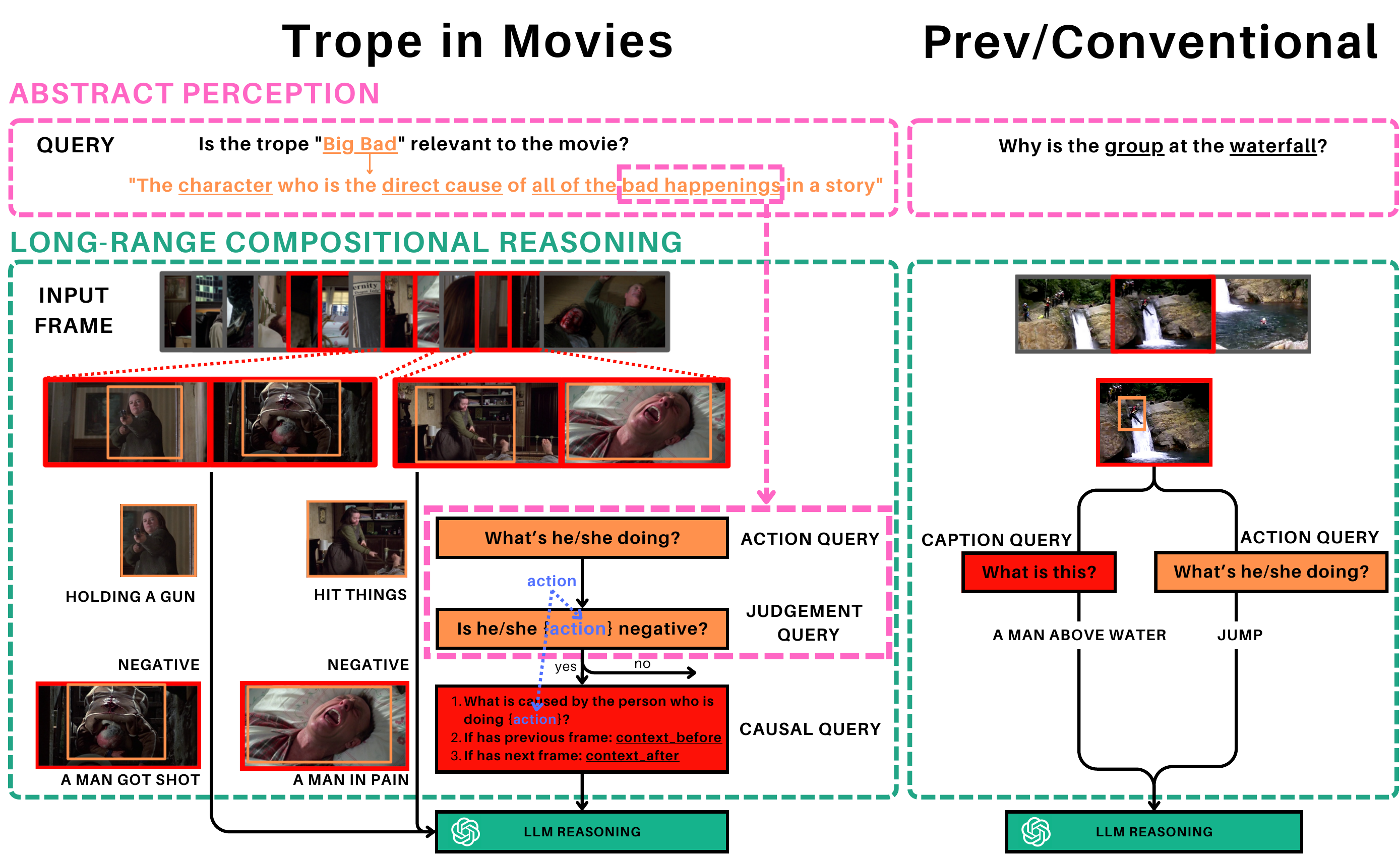}
  \caption{
  %Comparison of challenges between  \datasetabbr{} and existing datasets.
  Compared to previous datasets like NExT-QA \cite{nextqaxiao2021next}, \datasetfull{} (\datasetabbr{}) introduces the challenges of \textbf{\challengeii{}} (upper box) and \textbf{\challengeiii{}} (lower box), offering a robust framework for evaluating and developing LLM-based methods. The blue text (action) indicates that the answer to the action query will affect the input of the judgment query and causal query, which means decomposing these complex elements necessitates multiple, nested queries that are interdependent.
  }
\label{fig:tim_fig1}
\end{figure}

Large Language Models (LLMs) \cite{brown2020languagegpt3,gpt4,li2023llama,team2023gemini} have not only dominated Natural Language Processing but also extended their reach into Computer Vision (CV) reasoning tasks. Leveraging LLMs as their foundation, various video reasoning models have been introduced.
%to harness the reasoning capabilities of LLMs and address the challenges posed by visual language gaps. 
\textit{Captioner-Reasoner (C-R)} \cite{lin2021vx2text,wang2022language,lin2023towards,lsscaptionerreasonerchung2023long,llovicaptionerreasonerzhang2023simple} leverages visual language models (VLMs) to tokenize visual inputs into language tokens to feed into LLMs. While there may be potential information loss during captioning, C-R achieves remarkable performance on various video reasoning tasks such as NexT-QA \cite{nextqaxiao2021next}. 
\textit{Large Multimodal Model Instruction Fine-tuning (LMM-IF)} \cite{videollamazhang2023video,sevilayu2023self,llamavidli2023llama} aligns visual inputs to LLMs' token space using projection layers, thereby avoiding information loss during captioning. 
\textit{Visual Programming (VP)} \cite{visproggupta2023visual,suris2023vipergpt} harnesses LLMs to generate programs that call visual perception modules and integrate their outputs. In contrast to the C-R and LMM-IF approaches, VP facilitates ``System 2 style'' stepwise reasoning~\cite{evans2003two}. It demonstrates the capability to address complex reasoning tasks that require external knowledge or commonsense such as \cite{hudson2019gqa,okvqamarino2019ok,nextqaxiao2021next} in a stepwise and interpretable manner. 
While LLM-based methods demonstrate significant performance on existing benchmarks 
%and show promise for addressing future challenges
, several critical aspects remain underexplored in current models and datasets, as shown in Figure \ref{fig:tim_fig1}. 
First, \textbf{\challengeii{}}: While most queries in existing datasets target concrete elements like actions, objects, or attributes—easily captured by vision models—abstract concepts such as emotion, motivation, humor, and judgment remain obscure and continue to challenge advanced VLMs. Second, \textbf{\challengeiii{}}: Traditional datasets often assume that context and queries are straightforward, suitable for sparse sampling and simple decomposition. However, the reality is that contexts can span hour-long videos with thousands of frames, and queries may involve a wide range of complex elements. Decomposing these complex elements necessitates multiple, nested queries that are interdependent.
%Although the context and queries are typically sparse and can be decomposed into straightforward, basic elements, the context may involve hour-long videos containing thousands of frames, and the queries could encompass diverse and complex elements. 
%First, \textbf{\challengei{}}: Existing datasets primarily focus on minute-long videos, which can often be effectively addressed with sparse sampling to reduce the computational load from processing numerous frames. Second, \textbf{\challengeii{}}: The queries predominantly focus on concrete elements such as actions, objects, or attributes, which can be readily captured by vision models. Third, \textbf{\challengeiii{}}: The query can be readily decomposed into base elements with clean and straightforward structures. 
Consequently, these prevailing approaches may overlook the nuanced interplay between visual cues and complex linguistic structures in longer, more dynamic sequences. 
%Furthermore, existing methodologies struggle to demonstrate their capability to handle ambiguous or abstract queries that require deeper semantic understanding and contextual interpretation.

To evaluate these capabilities, we introduce a novel dataset, \textbf{\datasetfull{} (\datasetabbr{})}, designed to rigorously test existing and future LLM-based video reasoning models against the challenges identified as \challengeii{} and \challengeiii{}. 
On \datasetabbr{}, the model has to determine whether a \textit{trope} is present. \textit{Tropes} are commonly employed narrative devices that enable storytellers to craft situations easily recognizable to audiences \cite{timoschang2021situation}.
For instance, the trope ``Big Bad'' refers to an antagonist who is responsible for all the negative events in a story and drives the plot forward. 
%“Heroic Sacrifice” occurs when a character saves others from harm at the cost of their own life. %abstraction...challenges
Recognizing such a trope within varied narrative contexts demands \challengeii{} and \challengeiii{} from a machine learning model. 
\challengeii{} allows the model to identify the essential characteristics of the ``Big Bad'' trope beyond specific instances, encompassing a range of characters, judgments, motivations, and actions that fit the trope's broad definition. 
\challengeiii{} enables the model to process thousands of frames and decompose the concept of ``Big Bad'' into aspects such as evil characteristics, negative judgments, and the causation of terrible events. It also helps locate the relevant frames from among thousands to determine whether the trope is present. 
%Scalability is required as the model must process potentially vast narratives or series where the antagonist's influence unfolds over many episodes or chapters. 
%Compositional reasoning enables the model to understand how different narrative elements—such as character actions, plot developments, and thematic elements—interact to categorize a character as the ``Big Bad.'' Together, these capabilities ensure that the model can accurately interpret and predict narrative structures in complex and dynamically evolving stories.

We conducted comprehensive experiments on \datasetabbr{} using state-of-the-art (SOTA) LLM-based methods. These SOTA methods achieved a maximum F1 score of 25, only marginally surpassing the random baseline and significantly lagging behind human performance (65 F1 \cite{timoschang2021situation}). 
Even Gemini-1.5 \cite{team2023gemini}, which is known for multimodal long-context abilities, only reaches 40 F1. 
This underscores that advanced LLM-based video reasoning methods, including C-R \cite{llovicaptionerreasonerzhang2023simple}, LMM-IF \cite{sevilayu2023self,llamavidli2023llama}, and VP \cite{suris2023vipergpt}, struggle with the \challengeii{} and \challengeiii{} challenges presented by \datasetabbr{}. Consequently, \datasetabbr{} could serve as an effective testbed for further developing and evaluating future LLMs. Additionally, we have enhanced ViperGPT \cite{suris2023vipergpt} by introducing a \methodfull{} (\methodabbr{}) that fosters role awareness and a \promptfull{} (\promptabbr{}) that decouples context from query during reasoning, which improved the F1 score of base ViperGPT by 15 points. However, the performance still lags significantly behind human benchmarks (40 vs. 65 F1), indicating substantial room for improvement.

We conducted a comprehensive ablation study on \methodabbr{} to explore the impact of \challengeii{} and \challengeiii{}. Our findings reveal that \datasetabbr{}: (1) requires a higher number of frames to achieve optimal performance, with a noticeable decrease (-2.8 F1) when sparse sampling methods—commonly employed in many models—are used; (2) sees a significant improvement (+4.5 F1) with the adoption of advanced VLM (replace BLIP-2 \cite{li2023blip2} with Gemini \cite{team2023gemini}) that bolster abstraction; and (3) shows that GPT-4 \cite{gpt4} performs only marginally better (by 0.17 F1) than GPT-3.5.

To more accurately quantify the challenges of \challengeii{} and \challengeiii{} in datasets, we examine the abstract syntax tree (AST) of code generated by (VP). 
We propose a novel framework, \evaluationfull{} (\evaluationabbr{}), which is AST-based, to evaluate the levels of \challengeii{} and \challengeiii{}. \evaluationabbr{} quantifies \challengeii{} by counting VLM calls and token lengths, and examines \challengeiii{} through the nodes and edges of the AST. \evaluationabbr{} reveals that \datasetabbr{} necessitates code with higher \challengeii{} and \challengeiii{}. It also provides a useful tool for quantifying challenges in video reasoning for future tasks.

The contributions of this work are summarized as follows:
\begin{itemize}
\item We introduce a novel dataset, \textbf{\datasetfull{} (\datasetabbr{})}, designed to assess the \challengeii{} and \challengeiii{} aspects of video reasoning.
\item We demonstrate that SOTA LLM-based video reasoning methods, including Captioner-Reasoner \cite{llovicaptionerreasonerzhang2023simple}, Large Multimodal Model Instruction Fine-tuning \cite{sevilayu2023self,llamavidli2023llama}, and Visual Programming \cite{suris2023vipergpt}, face \challengeii{} and \challengeiii{} challenges in effectively tackling \datasetabbr{}.
\item We enhanced Viper \cite{suris2023vipergpt} by introducing \methodabbr{} and \promptabbr{}. These enhancements respectively enable role awareness and the decoupling of context from the query, facilitating progressive reasoning. This approach improved the F1 score by 15 points, marking a significant step toward reaching human-level performance (40 vs. 65 F1).
\item We have established a protocol, \evaluationfull{} (\evaluationabbr{}), which utilizes the abstract syntax tree (AST) of generated code to evaluate the levels of \challengeii{} and \challengeiii{} in datasets. \evaluationabbr{} not only highlights the unique challenges presented by \datasetabbr{} compared to previous models but also provides a valuable tool for future research to analyze datasets.
\end{itemize}

%inferior performance (human-written code, human)

\section{Related Work}\label{sec:2:rw}
\subsection{Comparison to Existing Tasks}
\datasetabbr{} presents a unique challenge in video reasoning, requiring \challengeii{} and \challengeiii{}. Most existing benchmarks primarily focus on identifying specific objects, actions, or attributes in short video clips \cite{videoqazeng2017leveraging,msvdqaxu2017video,anetqayu2019activitynet}. TVQA \cite{lei2018tvqa,lei2020tvqa+}, which leverages TV series similar to the movies used in our benchmark, creates a dataset centered on temporal relations. More recent datasets have advanced further to include causal relations \cite{nextqaxiao2021next,starbcsstar,causalvidqali2022representation} and incorporate external knowledge \cite{okvqamarino2019ok}. While these tasks pose challenges for conventional end-to-end video QA models, LLM-based models significantly enhance performance in a training-free manner by tokenizing inputs and incorporating commonsense knowledge from LLMs. For instance, training-free LLM-based methods \cite{llovicaptionerreasonerzhang2023simple,sevilayu2023self,suris2023vipergpt} outperform previous supervised models \cite{xiao2022video,vgtxiao2022video} that were specifically trained for Video QA tasks. 
While several datasets \cite{zhang2023movqa, mangalam2024egoschema} attempt to assess the model's capability to handle long-range videos, they do not incorporate the same levels of \challengeii{} and \challengeiii{}. TrUMAn \cite{su2021truman} is another dataset that uses tropes in video clips to evaluate machine learning models; however, it utilizes short clips featuring a single trope and does not involve the same depth of \challengeiii{}. Therefore, we are optimistic that \datasetabbr{} will further advance the development of LLM reasoning capabilities.

\subsection{Tropes in Movies}
Tropes are tools used in creative works and are leveraged for automatic content creation assistance \cite{smith2017harnessing,chou2023talestream}, or to serve as a testbed for evaluating the reasoning skills of machine learning models \cite{timoschang2021situation,su2021truman}. 
TiMoS \cite{timoschang2021situation} compiles movie synopses from the IMDb dataset and associates these with trope annotations from the TVTropes database. TiMoS serves as a benchmark to test NLP models and demonstrates that supervised models (e.g., BERT \cite{devlin2018bert}) struggle to reason about tropes in movie synopses. Since these models access human-written synopses instead of the movie, simplifying the challenge of understanding visual inputs. In contrast, TrUMAn \cite{su2021truman} utilizes video clips annotated with tropes from TVTropes to create a video trope reasoning dataset. However, reasoning from short clips is considerably simpler than from full movies. \datasetabbr{} utilizes a subset of the TiMoS dataset and associates it with movies collected from the MovieNet dataset \cite{huang2020movienet}, enabling the evaluation of video reasoning capabilities with long videos.

\section{Trope in Movies (TiM) Dataset}\label{sec:3:dataset}
\paragraph{Overview}
%We introduce the novel dataset Tropes in Movies (\datasetabbr{}) , which 
\datasetabbr{} comprises (1) 684 movies, each annotated with per-shot keyframes, subtitles, and trope labels, and (2) 95 trope identification queries accompanied by their definitions. The \datasetabbr{} dataset is specifically designed to pose more demanding and intricate reasoning tasks in video analysis, particularly focusing on extended content such as movies. The homepage of the \datasetabbr{} dataset\footnote{https://ander1119.github.io/TiM/} offers a download link for the \datasetabbr{} data along with detailed explanations of the annotations. Additionally, we have provided a pre-processing script for our baseline models in Section~\ref{sec:4:exps} to facilitate reproduction of our experimental results.
%By leveraging trope recognition, this dataset confronts the complexities of plot evolution, character interactions, and situational contexts that typify cinematic narratives. Movies, with their layered storytelling and intricate character dynamics, serve as an excellent medium for evaluating sophisticated video question answering (video QA) models. 
% The TiM dataset not only enriches the field by utilizing trope identification as a proxy but also pushes the boundaries of current datasets that lack depth in understanding the sophisticated elements of film plots and character relationships.
\paragraph{Trope}
% \begin{figure}
%   % \centering
%   \includegraphics[width=.5\textwidth]{figures/trope_wordcloud.pdf}
%   \caption{Word cloud of trope occurrences in Fullset, size of the tropes in proportion to their frequency in Fullset and color of the tropes correspond to the category they belongs
%   }
% \label{fig:trope_wordcloud}
% \end{figure}

\begin{wrapfigure}{r}{0.5\textwidth}
  \centering
  \includegraphics[width=.48\textwidth]{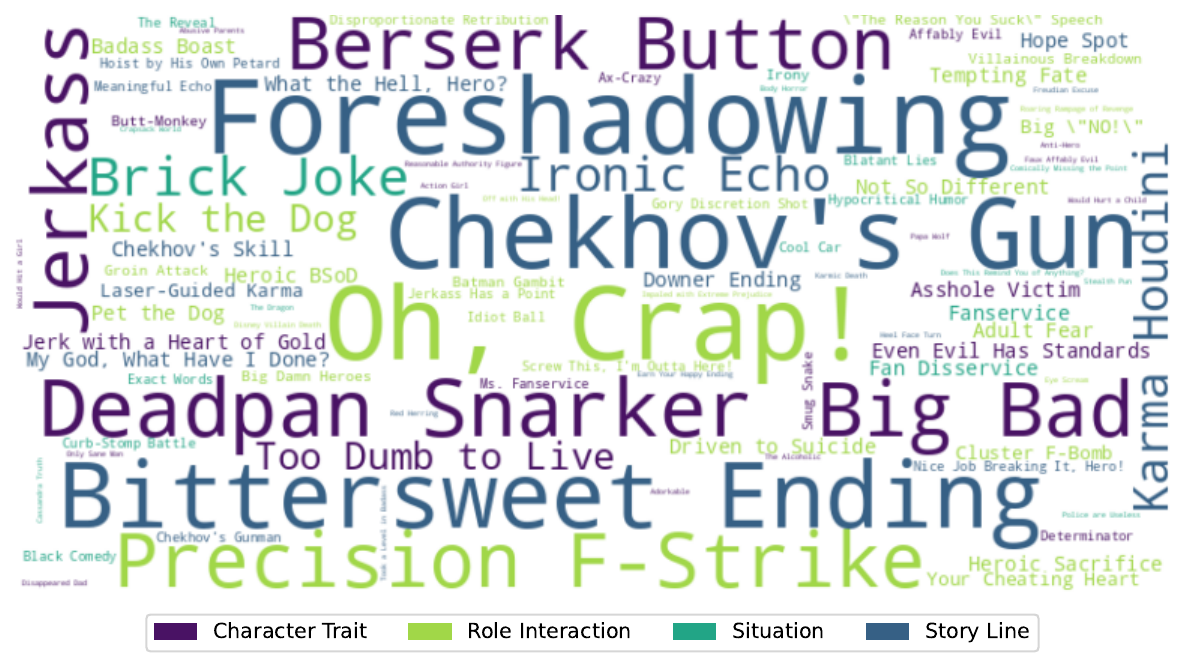}
  \caption{Word cloud of trope occurrences in Fullset, size of the tropes in proportion to their frequency in Fullset and color of the tropes correspond to the category they belongs
  }
\label{fig:trope_wordcloud}
\end{wrapfigure}
Considering the broad diversity of tropes, we utilize a set of 95 tropes categorized into four groups as introduced by TiMoS \cite{timoschang2021situation}, depicted in Figure \ref{fig:trope_wordcloud}. Subsequent research could explore expanding the dataset by incorporating additional tropes. 
%Given the extensive diversity of tropes, we adopt the classification from \cite{timoschang2021situation} for both understanding and analyzing tropes in experiments, as detailed in Section~\ref{sec:4:exps}. 
%The categories used are Character Traits, Role Interaction, Situation, and Storyline. Character Traits focus on the individual strengths and personalities of characters, illuminating how these attributes affect their behavior and interactions within the story. Role Interaction examines the dynamics between characters, especially how their relationships and behaviors influence the film’s development. Situation refers to specific scene-level scenarios that encapsulate abstract concepts and emotional dynamics, often driving the plot forward. Lastly, Storyline encompasses the overall narrative structure, guiding the flow and thematic elements throughout the film, typically running its entire length. Together, these categories provide a comprehensive framework for dissecting the complex interplay of tropes within cinematic narratives.
The categories used are Character Traits, Role Interaction, Situation, and Storyline. Character Traits analyze individual strengths and personalities, showing their impact on behavior and interactions within the story. Role Interaction explores the dynamics between characters and their influence on the film's development. Situation covers specific scene-level scenarios that drive the plot with abstract concepts and emotional dynamics. Storyline focuses on the overall narrative structure, guiding the flow and thematic elements throughout the film. Together, these categories offer a comprehensive framework for analyzing the complex interplay of tropes in cinematic narratives.

\paragraph{Task Definition}
%We transform the multi-label classification problem \cite{timoschang2021situation} into a binary classification query in order to simplify the original task and shift the focus more on complex reasoning in long-form video for a single trope. Future work might want to explore more challenging multi-label task.
We formulate the task considered here as binary classification: $y = f(\text{movie}, \text{trope})$, where \( y \in \{\text{True}, \text{False}\} \) indicates whether a given trope is present in the movie.
This simplifies the task and enhances the focus on complex reasoning for single tropes in movies.
Future research could consider revisiting the more challenging multi-label tasks~\cite{timoschang2021situation}. 
%We convert the multi-label classification problem \cite{timoschang2021situation} into a binary classification query to simplify the task and enhance focus on complex reasoning for single tropes in movies: $y = f(\text{movie}, \text{trope})$, where \( y \in \{\text{True}, \text{False}\} \) indicates whether a given trope is present in the movie. Future research could consider revisiting the more challenging multi-label tasks. 
%The task processes input frames, subtitles, and a defined trope, and determines whether the trope is present in the movie.

%The model would solely tasked with determining whether a movie contains a specific trope or not. This modification aims to enhance the precision of trope recognition and facilitate deeper analysis of cinematic content.

\paragraph{Evaluation}
We have selected the micro F1 score as the primary metric for global comparison within the chosen set in \datasetabbr{}. 
%Additionally, we present the micro F1 score for each trope category to compare performance across different types of tropes and to observe overall trends. This approach allows us to assess the model's effectiveness in accurately identifying each specific trope while providing a comprehensive view of its generalizability across various cinematic themes.
\begin{wraptable}{r}{0.6\textwidth}
 \caption{Comparison between different experiment setups.}
    \centering
    \begin{tabular}{l|ccccc}
    \toprule
    \multirow{2}{*}{Setting} & \multirow{2}{*}{Movies} & \multirow{2}{*}{Frames} & \multicolumn{2}{c}{Subtitles} & \multirow{2}{*}{Tropes} \\
                                &         &        & Line     & Char & \\
    \midrule
    Fullset                     & 684      & 1545.7  & -        & -         & 11.91 \\
    \midrule
    VDset                & 246      & 1585.9  & 1587.4   & 56k   & 13.38 \\
    Mainset              & 50       & 1699.6  & 1822.2   & 65k   &  6.08 \\
    \bottomrule
    \end{tabular}
    \label{tab:movie_analysis}
\end{wraptable}
\paragraph{Data Collection}
We sourced trope occurrences in movies from the TiMoS dataset \cite{timoschang2021situation}, originally compiled from the TVTropes database. Movie frames and subtitles were gathered from the MovieNet dataset \cite{huang2020movienet}. We aligned the movies with their corresponding tropes using their IMDb IDs. Future research could extend this dataset by collecting more movies.
%Initially, we utilized the 95 tropes and their definitions from the TiMoS dataset \cite{timoschang2021situation}, where the tropes are predominantly sourced from a Wikipedia-style database. Subsequently, we aligned the movies in the TiMoS dataset with those in MovieNet \cite{huang2020movienet} to identify overlapping entries. For the movies within this overlap, we extracted keyframes from MovieNet and also extracted subtitles from its annotations as different modality input source. Each movie was further annotated with trope labels derived from TiMoS dataset.

\paragraph{Data Statistics}
% The table\ref{tab:movie_analysis} outlines a comparative analysis of different experiment setups of \datasetabbr{}. The Fullset is the whole \datasetabbr{} dataset, while VDset and Mainset are subset of Fullset. Each set offering unique characteristics in terms of the number of movies, frames, subtitles, and tropes.
This benchmark is tailored for LLM-based methods, utilizing the entire dataset as the test set. Supervised learning evaluations are conducted using 5-fold cross-validation. To accommodate the absence of some subtitles in the MovieNet dataset, we offer the \textit{VDset}, which includes subtitles. Additionally, the \textit{Mainset}—a subset of 50 movies—is provided for more detailed analysis as experiments may require additional time or resources. Table \ref{tab:movie_analysis} presents a comparative analysis of different experimental setups.

%The table \ref{tab:movie_analysis} presents a comparative analysis of different experimental setups within the \datasetabbr{} dataset. The \textit{Fullset} represents the entirety of the \datasetabbr{} dataset, while the \textit{VDset} and \textit{Mainset} are specific subsets of the \textit{Fullset}. Each set exhibits unique characteristics in terms of the number of movies, frames, subtitles (both in lines and characters), and tropes. This segmentation facilitates a detailed evaluation of the dataset's diversity and complexity in various configurations.

%\paragraph{Accessibility}
%The homepage of the \datasetabbr{} dataset\footnote{https://ander1119.github.io/TiM/} offers a download link for the \datasetabbr{} data along with detailed explanations of the annotations. Additionally, we have provided a pre-processing script for our baseline models \ref{sec:4:exps} to facilitate reproduction of our experimental results.
%\subsection{Potential Extensions}

\section{Experiments}\label{sec:4:exps}

\subsection{Baselines}
%To assess the modern LLM-based approaches against the challenges presented in \datasetabbr{}, we evaluate \datasetabbr{} on three SOTA LLM-based approaches.
%selected one to two methods from each category: Captioner-Reasoner (C-R)), Large Multimodal Model Instruction Fine-tuning (LMM-IF), and Visual Programming (VP). 

\paragraph{Captioner-Reasoner} We tested LLoVi \cite{llovizhang2023simple}, which addresses video reasoning by tokenizing frames using VLMs such as BLIP-2 \cite{li2023blip2}. This efficient approach allowed LLoVi to achieve an accuracy of 67.7 on NExT-QA \cite{nextqaxiao2021next}. Given its success, LLoVi shows potential for handling more complex, long-range video QA tasks by effectively summarizing captions.

\paragraph{Large Multimodal Model Instruction Fine-tuning} SEVILA \cite{sevilayu2023self} introduces a two-stage pipeline that utilizes fine-tuned large multimodal models to localize keyframes and apply reasoning to selected frames, achieving an accuracy of 73.8 on NExT-QA with only 4 frames used for sparse sampling as inputs. Considering that \datasetabbr{} might require more input frames, we also incorporate LLaMA-VID \cite{llamavidli2023llama}, which adopts a different strategy by projecting frames into two tokens to efficiently handle long-range video inputs.

\paragraph{Visual Programming} ViperGPT \cite{suris2023vipergpt} leverages LLMs as a code generator that dynamically allocates VLMs and vision models, such as object detection, to progressively derive reasoning results. Although Viper may not always outperform LLoVi and SEVILA in terms of performance, it offers superior interpretability because the generated code illustrates how LLMs decompose tasks and perform stepwise reasoning.

% \paragraph{LLaMa-VID} LLaMa-VID \cite{li2023llama} address the computation burdens from processing long video by represent each frame with two token. The strategy make LLM preceiving critical information and also enlarge the context length from long video.

\paragraph{Gemini 1.5} To assess the limits of machine learning models, we tested Gemini 1.5 \cite{team2023gemini}, a trillion-scale model that significantly surpasses the size of previously mentioned models. This serves as a benchmark for future research.

\subsection{Proposed Method}
In our initial approach to \datasetabbr{}, we enhanced Viper \cite{suris2023vipergpt} with two novel features designed to address \challengeii{} and \challengeiii{} respectively.

\paragraph{\methodfull{} (\methodabbr{})} Previous datasets have primarily focused on short, simple clips rather than movies featuring numerous characters with rich interactions. Consequently, the original Viper design lacked tools specifically aimed at role identification. \methodabbr{} augments Viper by providing a face detection tool with examples in the prompts. \methodabbr{} enhances the fine-grained understanding of the ``human'' object to address \challengeii{}\footnote{Implementation details in Appendix \ref{subsec:a:fevori}}.

\paragraph{\promptfull{} (\promptabbr{})} Viper \cite{suris2023vipergpt} processes NExT-QA \cite{nextqaxiao2021next} by temporally locating frames or objects and querying the VLM about them. This approach struggles with \datasetabbr{} due to the intricate narratives of movies and the complex definitions of tropes. \promptabbr{} addresses the \challengeiii{} challenge by progressively decomposing the narrative context and trope query. It systematically checks if the extracted context matches each dimension of a trope through the generated program\footnote{Implementation details in Appendix \ref{subsec:a:conquer}}.

\begin{table}[]
\caption{State-of-the-art performance on TiM. everyshot: the model takes one frame per shot. SeViLA\textsuperscript{†}: SeViLA that uses the zero-shot localizer. 120$\rightarrow{}$ 16: SeViLA localizer selects 16 keyframes from 120 frames. 16\tiny{(SeViLA)} : Viper uses 16 frames selected by SeViLA localizer. FEVoRI\textsuperscript{*}: evaluate on \textit{Mainset}. Human: human evaluation result from \cite{timoschang2021situation}. we select Mainset as multi-modality setting for fair comparison}
    \centering
    \scalebox{0.85}{
        \begin{tabular}{@{}cllccccccc@{}}
        \toprule
            \multirow{3}{*}{\textbf{Modality}} & \multirow{3}{*}{\textbf{Method}} & \multirow{3}{*}{\textbf{\# Frames}} & \multirow{3}{*}{\textbf{Pre.}} & \multirow{3}{*}{\textbf{Rec.}}  & \multirow{3}{*}{\textbf{F1}}   &  \multicolumn{4}{c}{\textbf{Category F1}} \\
            \addlinespace[2pt]\cmidrule(lr){7-10}
             % \\
             & & & & & &  \textbf{CT} & \textbf{RI} & \textbf{ST} & \textbf{SL}  \\
        \midrule
            \multirow{7}{*}{V(Fullset)}
            &Random                                   & -                         & 12.24 & 48.48 & 19.54 & 19.23 & 19.99 & 17.37 & 23.37 \\ 
            &LLoVi \cite{llovizhang2023simple}        & everyshot                 & 20.47 & 17.67 & 18.97 & 13.46 & 16,67 & 15.22 & 25.58 \\
            &SeViLA\textsuperscript{†} \cite{sevilayu2023self}     & 120$\rightarrow{}$16        & 12.35 & 96.71 & 21.90 & 25.12 & 19.02 & 22.38 & 20.96 \\
            &SeViLA \cite{sevilayu2023self}     & 120$\rightarrow{}$16        & 15.29 & 51.75 & 23.61 & 23.46 & 23.43 & 17.81 & 27.58 \\
            &Viper \cite{surismenon2023vipergpt}      & 16 \tiny{(SeViLA\textsuperscript{†})}            & 13.26 & 67.33 & 22.15 & 21.58 & 22.63 & 19.92 & 24.60 \\
            &Viper \cite{surismenon2023vipergpt}      & 16 \tiny{(SeViLA)}        & 14.09 & 68.70 & 23.39 & 21.41 & 24.62 & 20.90 & 26.85 \\
            &FEVoRI\textsuperscript{*}                & 120                       & 27.07 & 32.32 & 29.42 & 12.36 & 22.75 & 35.62 & 48.78 \\
            &\textcolor{gray}{Gemini 1.5 \cite{team2023gemini}}& 120 &\textcolor{gray}{38.37} & \textcolor{gray}{34.42} & \textcolor{gray}{40.74} & \textcolor{gray}{40.45} & \textcolor{gray}{38.79} & \textcolor{gray}{38.55} & \textcolor{gray}{45.11} \\
        \midrule
            \multirow{7}{*}{V+D(Mainset\tablefootnote{Performance difference between Mainset and VDset in Appendix \ref{subsec:a:vd_main_diff}})}
            &Random                                   & -                         & 14.14 & 50.08 & 22.06 & 20.26 & 21.24 & 19.50 & 23.92 \\
            &LLoVi \cite{llovizhang2023simple}        & everyshot                 & 31.35 & 17.21 & 18.78 & 20.20 & 24.40 & 35.95 & 40.63 \\
            &SeViLA\textsuperscript{†} \cite{sevilayu2023self} & 120$\rightarrow{}$16 & 17.30 & 89.33 & 28.98 & 22.64 & 24.76 & 32.83 & 35.79 \\
            &SeViLA \cite{sevilayu2023self}      & 120$\rightarrow{}$16               & 22.98 & 58.18 & 28.54 & 28.92 & 25.00 & 37.50 & 42.86 \\
            &LLaMA-VID \cite{li2023llama} {}          & 240                       & 15.56 & 90.12 & 26.53 & 25.72 & 24.60 & 28.31 & 38.15 \\
            &Viper \cite{surismenon2023vipergpt}      & 16 \tiny{(SeViLA\textsuperscript{†})} & 14.58 & 37.87 & 21.05 & 18.15 & 14.35 & 20.58 & 31.56 \\
            &Viper \cite{surismenon2023vipergpt}      & 16 \tiny{(SeViLA)}        & 14.38 & 38.79 & 20.98 & 24.39 & 15.22 & 18.02 & 24.76 \\
            &Viper \cite{surismenon2023vipergpt}      & 120                       & 27.78 & 21.74 & 24.39 & 22.91 & 19.59 & 40.43 & 48.78 \\
            &FEVoRI                                   & 120                       & 27.88 & 39.80 & 32.79 & 30.52 & 29.55 & \textbf{42.42} & 49.67 \\
            &FEVoRI+ConQueR                           & 120                       & 32.11 & 51.28 & \textbf{39.64} & \textbf{42.80} & \textbf{34.48} & 39.78 & 
            \textbf{55.17} \\
        \midrule
            Synopses& Human \cite{timoschang2021situation}    & -                         & 65.77 & 63.98 & 64.87 & -     & -     & -     & -     \\
        \bottomrule
        \end{tabular}
    }
    \vspace{5pt}
    \label{tab:all_exp}
\end{table}

\begin{table}[]
    \centering
    \caption{Ablation study on \methodabbr{} framework on \datasetabbr{} \textit{Mainset}.}
    \scalebox{0.75}{
        \begin{tabular}{lcccc|lllcccc}
        \toprule
            & \multirow{3}{*}{\textbf{Modality}} & \multirow{3}{*}{\textbf{\# Frames}} & \multirow{3}{*}{\textbf{VLM}} & \multirow{3}{*}{\textbf{Coder}} & \multirow{3}{*}{\textbf{Pre.}} & \multirow{3}{*}{\textbf{Rec.}}  & \multirow{3}{*}{\textbf{F1}}   &  \multicolumn{4}{c}{\textbf{Category F1}} \\
                \addlinespace[2pt]\cmidrule(lr){9-12}
            & & & & & & & & \textbf{CT} & \textbf{RI} & \textbf{ST} & \textbf{SL}  \\
        \toprule
              \methodabbr{} & V+D & 120 & BLIP-2 & GPT-4  & 27.88 & 39.80 & 32.79                                                     & 30.52 & 29.55 & 42.42 & 49.67 \\
        \midrule
                            & \textbf{V} & 120 & BLIP-2 & GPT-4 & 27.07 & 32.23 & 29.42 \tiny{(\textcolor{red}{(-3.37$\downarrow$)})}    & 12.36 & 22.75 & 35.62 & 48.00 \\ % & 28.51 \\% 52.95 \\
                            & V+D   & \textbf{everyshot} & BLIP-2 & GPT-4 & 27.27 & 46.15 & 34.29 \tiny{(\textcolor{codegreen}{(+1.50$\uparrow$)})}   & 33.30 & 30.12 & 44.68 & 50.00 \\ % & 25.44 \\% 55.48 \\
                            & V+D & \textbf{16} & BLIP-2 & GPT-4 & 25.71 & 40.72 & 31.52 \tiny{(\textcolor{red}{(-1.27$\downarrow$)})}    & 23.74 & 25.56 & 38.83 & 47.54 \\ % & 26.37 \\% 49.54 \\
                            & V+D & 120 & \textbf{Gemini} & GPT-4 & 29.37 & 51.15 & 37.31 \tiny{(\textcolor{codegreen}{(+4.52$\uparrow$)})}    & 28.71 & 29.49 & 47.17 & 53.23 \\ % & 33.61 \\ % 55.76 \\
                            & V+D & 120 & BLIP-2 & \textbf{GPT-3.5} & 30.16 & 35.52 & 32.62 \tiny{(\textcolor{red}{(-0.17$\downarrow$)})}    & 27.18 & 30.34 & 39.56 & 38.65 \\ % & 22.75 \\ % 53.22 \\
        \bottomrule
        \end{tabular}
    }
    \label{tab:viper_exp}
\end{table}

\subsection{Setup}
Most models in our experiments are training-free, so the entire \datasetabbr{} dataset is used for testing. Additionally, we fine-tuned SeViLa on \datasetabbr{} in a supervised setting to evaluate its performance. For these experiments, we employed five-fold cross-validation and reported the average performance. 
For LLoVi \cite{llovicaptionerreasonerzhang2023simple}, we employ the standard prompt with BLIP-2 \cite{li2023blip2} to generate captions for each frame in every shot of \datasetabbr{}. This is followed by a multi-round summarizing process 
%, similar to the LLoVi \cite{llovicaptionerreasonerzhang2023simple} setting, 
to create a summary for each movie. These summaries, coupled with binary classification queries, are then inputted into an LLM to generate answers. In the multi-modality version, we enhance the summarization process by integrating captions with subtitles.
For SeViLA \cite{sevilayu2023self}, we use NExT-QA setting for both zero-shot and fine-tuned scenarios, enabling the Localizer to select 16 frames from a set of 120. These selected frames are used to address binary classification queries, with an enhancement in the multi-modality version where visual features are concatenated with subtitles before being processed by the LLM during both the localizer and answerer stages. 
For LLaMA-VID \cite{li2023llama}, we use long-video-tuning model, which was tuned with QA pair from MovieNet \cite{huang2020movienet}, to inference on binary classification query on \datasetabbr{}.
For Viper and our proposed method, we have adapted the NExT-QA prompt on \datasetabbr{} and use GPT-4 as the code generator. For \methodabbr{}, we have integrated additional face identification tools into the Viper API, specifically employing DeepFace \cite{deepfaceserengil2021lightface} for face recognition.
%by introducing a simplified code generation and reasoning prompt and aligning the number of frames and their indices according to the SeViLA zero-shot and fine-tuned results, maintaining consistency with the ViperGPT setting that uses GPT-3.5 for the code generator and reasoning module.

%To further understand the behavior of \methodabbr{}, we conducted a comprehensive ablation analysis. Most experiments were performed using 120 frames per video, rather than sparse sampling, to ensure that the model has access to detailed information. We selected an \textit{ablation set} comprising 50 movies and 20 tropes to ensure the experiments could be completed within a reasonable timeframe.

%For \methodabbr{}, we employ GPT-4 \cite{gpt4} for code generation, sampling 120 frames per video. DeepFace \cite{deepfaceserengil2021lightface} is utilized for face identification. We designed two prompts: a Simple Prompt, adapted from the one used for NExT-QA \cite{nextqaxiao2021next} queries, and a more complex one. For other vision models, we followed the same approach as with ViperGPT.

\subsection{Existing LLM-based state-of-the-arts cannot reason on \datasetabbr{}}
As shown in Table \ref{tab:all_exp}, all LLM-based baselines struggle with reasoning on \datasetabbr{}, achieving only random-level performance (first row of each block). This underscores that despite their significant achievements on various video reasoning benchmarks, state-of-the-art models are unable to overcome the challenges posed by \datasetabbr{}. Access to dialogues results in an F1 score improvement of 2-4 points. Captioner-Reasoner (LLoVi \cite{llovicaptionerreasonerzhang2023simple}) records lower F1 scores, indicating that the loss of information or the abstraction gap during video captioning may lead to subpar performance on \datasetabbr{}. LLoVi also achieves relatively better performance in the Storyline (SL) category, which focuses on the overall plot rather than on fine-grained details. LMM-IF methods, including SeViLa \cite{sevilayu2023self} which achieves significant performance on various benchmarks, and LLaMA-VID \cite{llamavidli2023llama} designed for long videos, often resort to blindly guessing ``yes.'' This approach typically results in high recall but poor precision. Fine-tuning SeViLa enhances performance through supervised learning. Viper \cite{suris2023vipergpt} achieves decent performance without resorting to blindly guessing ``yes,'' and maintains superior precision compared to SeViLa and LLaMA-VID. Gemini \cite{team2023gemini} achieves a 41 F1 score, surpassing all previously mentioned methods due to its larger scale of parameters and training data. However, it still significantly trails human performance \cite{timoschang2021situation}, scoring 41 compared to 65 F1. Comprehensive experiments show that SOTA LLMs still struggle to address challenges in \datasetabbr{}.

\subsection{\methodabbr{} Analysis}
\paragraph{\methodabbr{} significantly boosts the F1 score by 8.5.} Comparing Viper and FEVoRI in the second block of Table \ref{tab:all_exp}, our augmentation allows the VP LLM to understand character interactions, leading to substantial performance improvements, particularly in the CT (Character Traits) and RI (Role Interaction) categories, where fine-grained role interactions are crucial. Remarkably, even the visual-only \methodabbr{} outperforms the supervised SeViLA \cite{sevilayu2023self}, demonstrating the superior design of our methodology. 
As the performance gain is primarily from an 18.00 improvement in recall, while precision improves by only 0.1, we hypothesize that \methodabbr{} improves by effectively identifying more relevant cases.

\paragraph{\promptabbr{} further increases the F1 score by 6.9.} In the second block of Table \ref{tab:all_exp}
, comparing \methodabbr{} and \methodabbr{}+\promptabbr{}, the modified \promptabbr{} demonstrates how progressively decomposing the trope query and movie narrative context enhances understanding. 
%This approach, which might be unnecessary in previous tasks, is essential in \datasetabbr{}. 
\promptabbr{} also effectively filters key signals to extract crucial information from long-range videos. The performance improvement highlights promising directions for future work in addressing \challengeiii{}.
%demonstrates a more comprehensive approach in (1) \textbf{Causal Analysis}: Unlike the simple prompt, which tends to leave much of the causal reasoning to be inferred by the language model, the complex prompt proactively links character actions to plot events throughout the entire movie. This depth of analysis aids in making more informed decisions regarding the presence of Storyline and Role-Interaction tropes, even enhance \challengeiii{} ability of \methodabbr{} (2) \textbf{Complexity of Logical Construct and Queries}: Complex prompts utilize nuanced queries into character behavior, dialogue, and environmental interactions, providing a higher level of \challengeii{} and also a deeper exploration of the narrative context. In contrast, simple prompts focus primarily on clearly visible or explicitly stated elements, often overlooking subtler thematic nuances that are crucial for comprehensive trope analysis.
%Given that \datasetabbr{} demands a higher level of \challengeii{} and \challengeiii{}, we further boosted performance with complex prompt by 6.9, achieving an F1 score of 39.64. Although this is still well below the human performance benchmark of 64 F1 reported by \cite{timoschang2021situation}, it highlights promising avenues for future research in video reasoning.

\paragraph{A higher frame rate consistently outperforms sparse sampling.} Several tropes depend on fleeting, fine-grained details or a comprehensive understanding of the entire plot. We compared the density of frame sampling by evaluating every shot (approximately 1,000 frames) and 120 frames per video, alongside a sparse sampling method that uses only 16 frames per video, which is commonly used in many approaches. As shown in Table \ref{tab:viper_exp}, a higher frame rate leads to marginal yet consistent improvements, with every-shot sampling boosting the F1 score by 2.8 points across all categories. This indicates that while sparse sampling is efficient, it may compromise performance.

\paragraph{Enhancing VLM \challengeii{} improves performance by 4.5.} A core challenge of \datasetabbr{} is \challengeii{}, which involves tokenizing visual signals into coherent concepts. Table \ref{tab:viper_exp} shows that replacing BLIP-2 \cite{li2023blip2} with more advanced Gemini \cite{team2023gemini}, the F1 score is boosted by 4.5 as Gemini is capable to tackle more abstract queries.

\paragraph{GPT-4 shows a slight improvement over GPT-3.5 in program generation.} When replacing GPT-4 with GPT-3.5, the F1 score drops by 0.2, as shown in Table \ref{tab:viper_exp}, demonstrating that GPT-3.5 is capable to generate programs without \promptabbr{}. 

% \paragraph{A complex prompt further increases the F1 score by 6.9.} Given that \datasetabbr{} demands a higher level of \challengeiii{}, we hypothesize that more complex and sophisticated examples could enhance performance by effectively decomposing the structure of the trope. Through meticulous prompt engineering, we further boosted performance by 6.9, achieving an F1 score of 39.64. Although this is still well below the human performance benchmark of 64 F1 reported by \cite{timoschang2021situation}, it highlights promising avenues for future research in video reasoning.

% \input{tables/simple_complex}
%\subsection{\challengei{} Analysis}
%\paragraph{Modality}
%\paragraph{Frame Rate}
%\subsection{Abstraction Analysis}
%\paragraph{VLM Perception}
%\paragraph{Face Identification}
%\subsection{Composition Analysis}
%\paragraph{Coder}
%\paragraph{Reasoning Level}

%\section{Beyond Performance: Abstraction and Composition Level Analysis}\label{sec:5:abcd}
\section{\datasetabbr{} Requires More \challengeii{} and \challengeiii{}: Quantitative Evidence}\label{sec:5:abcd}
%Compare different tropes
%vs. NExT-QA, (Causal VidQA, STAR, ...)
\subsection{Abstract Syntax Tree (AST) for Visual Programming}
While Section \ref{sec:4:exps} effectively highlights the challenges of \challengeii{} and \challengeiii{} encountered with \datasetabbr{}, it is challenging to quantify the degree of the challenge. Hence, we propose an evaluation protocol to assess the degree of \challengeii{} and \challengeiii{}, leveraging the Abstract Syntax Tree (AST) of VP code. 
AST is a tree structure that represents the syntactic structure of a code snippet, thereby reflecting the complexity of the reasoning task addressed by VP. By decomposing VP code into an AST, we can assess the level of \challengeii{} by measuring VLM calls and the level of \challengeiii{} by analyzing the nodes and edges within the AST. More nodes indicate higher syntactic complexity, while more edges signify intricate relationships between code constructs. This detailed analysis provides insights into the sophistication of the logic used, making AST a valuable tool for evaluating the intricacies of VP tasks. Therefore, we propose a novel framework based on AST to analyze the \challengeii{} and \challengeiii{} level of a VP task based on generated code\footnote[5]{Implementation details in \ref{subsec:a:abcd}}.
%TODO: What is AST and why it measures A&C (high-level wise)
% \begin{table}[]\label{tab:abcd}
%     \centering
%     \begin{tabular}{|l|cc|cc|}
%     \toprule
%         & \multicolumn{2}{|c|}{Abstraction} & \multicolumn{2}{|c|}{Composition} \\
%         & VLM Calls & VLM Tokens & AST Nodes & AST Edges \\
%         %\multirow{5}{*}{V}
%         NExT-QA \cite{nextqaxiao2021next} & 1.59 & 9.49 & 102.09 & 146.32 \\
%         GQA \cite{hudson2019gqa} & 1.34 & 11.28 & 42.16 & 55.63 \\
%         OKVQA \cite{okvqamarino2019ok} & 1.66 & 12.04 & 42.50 & 58.46 \\
%         \datasetabbr{} (Simple Prompt) & 1.44 & 9.44 & 123.19 & 178.01 \\
%         \datasetabbr{} (Complex Prompt) & \textbf{1.84} & \textbf{14.09} & \textbf{141.81} & \textbf{205.06} \\
%         \bottomrule
%     \end{tabular}
%     \caption{\evaluationfull{} (\evaluationabbr{}) results of \datasetabbr{} and existing datasets. (Section \ref{sec:5:abcd})}
% \end{table}
\begin{table}[]
\caption{We propose an \evaluationfull{} (\evaluationabbr{}) to assess the levels of \challengeii{} and \challengeiii{} in a dataset, using code generated by VP. A higher number indicates greater complexity and challenge. (Section \ref{sec:5:abcd})}
    \centering
    \begin{tabular}{lcccc}
    \toprule
        \multirow{3}{*}{\textbf{Dataset}}          & \multicolumn{2}{c}{\textbf{\challengeii{}}} & \multicolumn{2}{c}{\textbf{\challengeiii{}}} \\
                                            \cmidrule(lr){2-3}                \cmidrule(lr){4-5}
                                          & \textbf{VLM Calls} & \textbf{VLM Tokens} & \textbf{AST Nodes} & \textbf{AST Edges} \\
        %\multirow{5}{*}{V}
        \midrule
        NExT-QA \cite{nextqaxiao2021next} & 1.60 & 11.15 & 102.09 & 146.32 \\
        GQA \cite{hudson2019gqa}          & 1.34 & 12.69 & 42.16 & 55.63 \\
        OKVQA \cite{okvqamarino2019ok}    & 1.66 & 13.75 & 42.50 & 58.46 \\
        \datasetabbr{} (w/o \promptabbr{})    & 1.77 & 14.11 & 123.19 & 178.01 \\
        \datasetabbr{} (w/ \promptabbr{})   & \textbf{1.97} & \textbf{20.67} & \textbf{141.81} & \textbf{205.06} \\
        \bottomrule
    \end{tabular}
    \vspace{5pt}
    \label{tab:abcd_tab}
\end{table}

\subsection{AST Based Code Diagnosis (ABCD)}
%\subsubsection{\challengeii{} Level Analysis}
\paragraph{\challengeii{} Level Analysis}
%\paragraph{VLM Calls} 
\textbf{VLM calls} serve as the primary interface for connecting visual inputs and transferring them to language representations. The frequency of VLM calls reflects the abstraction requirements for a visual programming task. 
%In the Viper API \cite{surismenon2023vipergpt}, which this work utilizes for visual programming, VLM calls are represented as \textit{simple\_query} calls in an AST tree.
%\paragraph{VLM Tokens} 
\textbf{VLM Tokens} indicate the complexity of VLM calls, which can vary from simple questions like "What is it doing?" to more complex and abstract inquiries such as "What is caused by the person doing {action}?" Facing the challenge of directly assessing the \challengeii{} level of a VLM query, we have developed a proxy method. This approach estimates the token length of a VLM call, based on the premise that more abstract concepts generally require a greater number of tokens for explanation in VLM models.

%weighted token?
%\paragraph{Simple Vision Calls}
%\paragraph{Total Vision Calls}
%\subsubsection{\challengeiii{} Level Analysis}
%\paragraph{AST Nodes}
\paragraph{\challengeiii{} Level Analysis}
\textbf{AST Nodes} represent a construct like statements, expressions, or operators, which when analyzed collectively through the count of nodes, provides a quantitative measure of the code's structural complexity. Essentially, each node encapsulates a specific element or operation in the code, and more nodes typically indicate more constructs and interactions. Therefore, a higher count of nodes in visual programming often indicates a more complex and intricate codebase, filled with numerous functional components and logical constructs, necessitated by tasks that require a higher level of \challengeiii{}. 
\textbf{AST Edges} denote the relationships between nodes, which are vital for understanding the structural and logical organization of code. Each edge connects nodes in a way that reflects the syntactic dependencies and execution order within the program, effectively mapping out the flow of control and data. A higher number of edges generally indicates a more complex interplay of these dependencies, suggesting more intricate code logic and increased interactions among the program's components. Thus, in VP, a dense network of AST edges usually points to sophisticated program constructs and a higher degree of \challengeiii{}, as tasks often necessitate nuanced combinations and sequences of operations to achieve desired functionalities and outcomes.
%\paragraph{AST Max Path}
%In an AST, nodes represent the syntactic constructs of the code, such as expressions, statements, and declarations, while edges represent the hierarchical and syntactic relationships between these constructs. In visual programming, more nodes in an AST typically indicate greater syntactic complexity or a more detailed breakdown of the code, while more edges signify more intricate relationships and dependencies between these syntactic elements.

%\subsection{VLM and Vision Model Calls}

\subsection{Results}\label{subsec:5:abcd_results}
%Table \ref{tab:abcd_tab} illustrates the levels of \challengeii{} and \challengeiii{} as assessed by \evaluationabbr{}. 
As shown in Table \ref{tab:abcd_tab}, it is clear that \datasetabbr{} requires a higher level of both \challengeii{} and \challengeiii{}, even without \promptabbr{}. 
Regarding \challengeii{}, \datasetabbr{} requires more VLM calls and a greater number of tokens to effectively process visual inputs from videos. As for \challengeiii{}, this results in a higher number of AST nodes and edges. 
Furthermore, adopting \promptabbr{} not only increases AST nodes and edges but also adds to the number of VLM calls and tokens. 
This analysis not only measures performance but also examines \challengeii{} and \challengeiii{} based on the complexity of the generated programs.

%\subsection{Abstraction Level Analysis}

%\include{figures/TiM_abstraction}
%Abstraction: 
%more VLM calls = high abstraction
%Complex VLM calls = high abstraction
%\subsection{Composition Level Analysis}
%\include{figures/TiM_composition}

%\include{figures/TiM_VLM_WC}
%Composition
%more LLM calls: high composition
%more LLM calls: high composition
%\subsection{Abstract Syntax Trees}
%Composition
%More assignments, nodes, edges, path length

\section{Conclusion}\label{sec:5}
We introduce a novel task, \datasetabbr{}, accompanied by a new dataset designed to test the challenges of \challengeii{} and \challengeiii{}. Our findings reveal that SOTA LLM-based methods such as Captioner-Reasoner, Large Multimodal Model Instruction Fine-tuning, and Visual Programming, lack the capabilities to meet these challenges effectively. To enhance performance, we have augmented the VP model \cite{suris2023vipergpt} with \methodabbr{} and \promptabbr{}, achieving a 15-point improvement in F1 score. Additionally, we propose a new protocol, \evaluationabbr{}, to assess the \challengeii{} and \challengeiii{} levels of datasets using code generated by VP. We believe that \datasetabbr{} could serve as a valuable testbed for the development and refinement of novel LLM-based video reasoning methods.

{
\small
\bibliographystyle{unsrtnat}
\bibliography{main}
}

\appendix

\section{Performance Comparison between Mainset and VDset}\label{subsec:a:vd_main_diff}
\begin{table}[h]
\caption{State-of-the-art performance on multi-modality settings (VDset and Mainset). everyshot: the model takes one frame per shot. SeViLA\textsuperscript{†}: SeViLA that uses the zero-shot localizer. 120{$\rightarrow$} 16: SeViLA localizer selects 16 keyframes from 120 frames. {16\tiny{(SeViLA)}} : Viper uses 16 frames selected by SeViLA localizer.}
    \centering
    \scalebox{0.9}{
        \begin{tabular}{@{}cllccccccc@{}}
        \toprule
            \multirow{3}{*}{\textbf{Modality}} & \multirow{3}{*}{\textbf{Method}} & \multirow{3}{*}{\textbf{\# Frames}} & \multirow{3}{*}{\textbf{Pre.}} & \multirow{3}{*}{\textbf{Rec.}}  & \multirow{3}{*}{\textbf{F1}}   &  \multicolumn{4}{c}{\textbf{Category F1}} \\
            \addlinespace[2pt]\cmidrule(lr){7-10}
             % \\
             & & & & & &  \textbf{CT} & \textbf{RI} & \textbf{ST} & \textbf{SL}  \\
        \midrule
            \multirow{7}{*}{VDset}
            &LLoVi \cite{llovizhang2023simple}        & everyshot                 & 19.23 & 21.73 & 20.40 & 20.85 & 24.87 & 19.49 & 31.62 \\
            &SeViLA\textsuperscript{†} \cite{sevilayu2023self}     & 120$\rightarrow$16       & 14.82 & 92.97 & 25.56 & 24.34 & 27.50 & 19.85 & 29.50 \\
            &SeViLA \cite{sevilayu2023self}     & 120$\rightarrow$16                & 16.32 & 65.21 & 26.11 & 26.08 & 28.89 & 18.29 & 28.57 \\
            &LLaMA-VID \cite{li2023llama} {}          & 240                      & 14.47 & 98.30 & 25.22 & 24.74 & 26.34 & 19.85 & 29.27 \\
            &Viper \cite{surismenon2023vipergpt}      & 16 \tiny{(SeViLA\textsuperscript{†})}            & 16.08 & 46.24 & 23.86 & 17.78 & 24.06 & 19.65 & 31.21 \\
            &Viper \cite{surismenon2023vipergpt}      & 16 \tiny{(SeViLA)}        & 16.48 & 44.41 & 24.04 & 21.01 & 27.30 & 19.72 & 27.79 \\
        \midrule
            \multirow{7}{*}{Mainset}
            &LLoVi \cite{llovizhang2023simple}        & everyshot                 & 31.35 & 17.21 & 18.78 & 20.20 & 24.40 & 35.95 & 40.63 \\
            &SeViLA\textsuperscript{†} \cite{sevilayu2023self} & 120$\rightarrow$16 & 17.30 & 89.33 & 28.98 & 22.64 & 24.76 & 32.83 & 35.79 \\
            &SeViLA \cite{sevilayu2023self}      & 120$\rightarrow$16               & 22.98 & 58.18 & 28.54 & 28.92 & 25.00 & 37.50 & 42.86 \\
            &LLaMA-VID \cite{li2023llama} {}          & 240                       & 15.56 & 90.12 & 26.53 & 25.72 & 24.60 & 28.31 & 38.15 \\
            &Viper \cite{surismenon2023vipergpt}      & 16 \tiny{(SeViLA\textsuperscript{†})} & 14.58 & 37.87 & 21.05 & 18.15 & 14.35 & 20.58 & 31.56 \\
            &Viper \cite{surismenon2023vipergpt}      & 16 \tiny{(SeViLA)}        & 14.38 & 38.79 & 20.98 & 24.39 & 15.22 & 18.02 & 24.76 \\
        \bottomrule
        \end{tabular}
    }
    \vspace{5pt}
    \label{tab:main_vd}
\end{table}
Table \ref{tab:main_vd} displays the performance difference between the Mainset and VDset for the baseline models we selected. The gap is relatively small, and for a fair comparison, we have chosen Mainset as the multi-modality setting in Table \ref{tab:all_exp}.

\section{Implementation Details}\label{sec:a:impl}

\subsection{\methodabbr{}}\label{subsec:a:fevori}
\lstset{
    backgroundcolor=\color{backcolour},  
    commentstyle=\color{codegreen},
    keywordstyle=\color{magenta},
    numberstyle=\tiny\color{codegray},
    stringstyle=\color{codepurple},
    basicstyle=\ttfamily\customsize,
    breakatwhitespace=false,         
    breaklines=false,                 
    captionpos=b,                    
    keepspaces=false,                 
    numbers=left,                    
    numbersep=1pt,                  
    showspaces=false,                
    showstringspaces=false,
    showtabs=false,                  
    tabsize=2
}

\begin{lstlisting}[language=Python, caption=FEVoRI ICL Example]
def execute_command(video, annotation, possible_answers, query):
    # Trope: Big Bad
    # Definition: The character who is the direct cause of all of the bad happenings in a story.
    # Thought Process:
    # 1. Frame Selection: Analyze each frame to identify key characters and their actions.
    # 2. Character Analysis: Identify the main antagonist and their actions throughout the video.
    # 3. Answer Selection: Determine if there is a single character causing most of the negative events.

    video_segment = VideoSegment(video, annotation)
    info = {
        "character_actions": {},
        "negative_impacts": {}
    }
    for i, frame in enumerate(video_segment.frame_iterator()):
        # Identify all characters in the frame
        for character in frame.find("person"):
            character_id = video_segment.face_identify(character)
            if character_id is None:
                continue
            # Query the action of the character in the frame
            action_query = frame.simple_query("What is this person doing?")
            # Check if the action has a negative impact
            negative_query = f"Does the action '{action_query}' have a negative impact?"
            has_negative_impact = frame.llm_query(negative_query, to_yesno=True)
            # Store character actions and their impacts
            if character_id not in info["character_actions"]:
                info["character_actions"][character_id] = []
            info["character_actions"][character_id].append(action_query)
            if "yes" in has_negative_impact.lower():
                if character_id not in info["negative_impacts"]:
                    info["negative_impacts"][character_id] = 0
                info["negative_impacts"][character_id] += 1

    # After collecting information, use it to determine the presence of the trope
    answer, reason = video_segment.select_answer(info, query, possible_answers)
    return answer, reason, info
\end{lstlisting}
We have integrated \textit{face\_identify}, which utilizes DeepFace \cite{deepfaceserengil2021lightface} to assign a unique ID to each character. As shown in Line 17, \methodabbr{} enhances fine-grained tokenization, extending beyond the generic object "human" to more effectively address \challengeii{}.

\subsection{\promptabbr{}}\label{subsec:a:conquer}
\lstset{
    backgroundcolor=\color{backcolour},  
    commentstyle=\color{codegreen},
    keywordstyle=\color{magenta},
    numberstyle=\tiny\color{codegray},
    stringstyle=\color{codepurple},
    basicstyle=\ttfamily\customsize,
    breakatwhitespace=false,         
    breaklines=false,                 
    captionpos=b,                    
    keepspaces=false,                 
    numbers=left,                    
    numbersep=1pt,                  
    showspaces=false,                
    showstringspaces=false,
    showtabs=false,                  
    tabsize=2
}

\begin{lstlisting}[language=Python, caption=ConQueR ICL Example]
def execute_command(video, annotation, possible_answers, query)->[str, str, dict]:
    # Trope: Big Bad
    # Definition: The character who is the direct cause of all of the bad happenings in a story.
    # Thought Process:
    # 1. Character Identification: Identify characters and track their actions across frames.
    # 2. Event Linking: Determine which negative events are directly caused by the actions of a character.
    # 3. Consistency Check: Check for consistency in the character's negative influence over the story arc.
    video_segment = VideoSegment(video, annotation)
    # Initialize a dictionary to store information collected during analysis
    info = {
        "happened bad events": {},
        "character infos": {}
    }
    for i, frame in enumerate(video_segment.frame_iterator()):
        for person in frame.find("person"):
            # identify the person in the frame
            person_id = video_segment.face_identify(person)
            if person_id is None:
                continue
            # query the character"s description and add into character_description
            if person_id not in info["character infos"]:
                descriptino_query = "Please describe his/her appearance in 10 words"
                character_description = person.simple_query(descriptino_query)
                info["character infos"][person_id] = {
                    "description": character_description,
                    "actions": {}
                }
            # query the character"s action in the frame
            action = person.simple_query("Please describe his/her action in the scene")
            info["character infos"][person_id]["actions"][f"{i} frame"] = action
        # check if there is any negative event happening in the scene
        check_negative_query = "Is there any negative event happening in the scene?"
        any_negative_event = frame.simple_query(check_negative_query, to_yesno=True)
        if "yes" in any_negative_event.lower():
            # query the negative events happening in the scene
            event = frame.simple_query("What's happening in the scene")
            info["happened bad events"][f"{i} frame"] = {
                "event": event,
                "potential cause": []
            }
            for pid, character_infos in info["character infos"].items():
                # check if the character is a potential cause of the negative event
                character_description = character_infos["description"]
                for prev_i in range(i, max(i-5, 0), -1):
                    prev_action = character_infos["actions"].get(f"{prev_i} frame", None)
                    if prev_action is not None:
                        person_query = f"Is person with '{character_description}' a potential cause of '{event}'?"
                        is_person_potential = frame.simple_query(person_query, to_yesno=True)
                        action_query = f"Is action '{prev_action}' a potential cause of '{event}'?"
                        is_action_potential = frame.simple_query(action_query, to_yesno=True)
                        if "yes" in is_person_potential.lower() or "yes" in is_action_potential:
                            info["happened bad events"][f"{i} frame"]["potential cause"].append(pid)
                        break
    # After collecting information, use it to determine the presence of the trope
    answer, reason = video_segment.select_answer(info, query, possible_answers)
    return answer, reason, info
\end{lstlisting}
\promptabbr{} enhances the model's ability to tackle \challengeiii{} by decomposing the movie narrative (context) and the trope (query). In this instance, \promptabbr{} systematically breaks down the identified characters and actions to align with the ``Big Bad'' trope query, as demonstrated in Lines 33, 36, 48, and 50.

\subsection{\evaluationabbr{}}\label{subsec:a:abcd}
We utilized all generated code from \datasetabbr{} and sampled 512 codes from NExT-QA \cite{nextqaxiao2021next}, OKVQA \cite{okvqamarino2019ok}, and GQA \cite{hudson2019gqa}. We constructed AST trees using the Python AST module and excluded codes that could not be parsed by AST (less than 3\%) from our analysis. For VLM token analysis, we used NLTK's word\_tokenize to split the VLM queries into tokens. The implementation details can be found in the repository.
\end{document}